\documentclass{article}

\usepackage{times}
\usepackage{graphicx} 
\usepackage{subfigure} 

\usepackage{natbib}

\usepackage{algorithm}
\usepackage{algpseudocode}

\usepackage{amsmath}
\usepackage{amssymb}
\usepackage{gensymb}

\newtheorem{definition}{Definition}

\usepackage[nohyperref,accepted]{icml2017}

\icmltitlerunning{Consistent feature attribution for tree ensembles} 

\begin{document} 

\twocolumn[
\icmltitle{Consistent feature attribution for tree ensembles}




\icmlauthor{Scott M. Lundberg}{slund1@cs.washington.edu}
\icmladdress{Paul G. Allen School of Computer Science,
            University of Washington, Seattle, WA 98105 USA}
\icmlauthor{Su-In Lee}{suinlee@cs.washington.edu}
\icmladdress{Paul G. Allen School of Computer Science and Department of Genome Sciences,  University of Washington, Seattle, WA 98105 USA}

\icmlkeywords{interpretable, machine learning, ICML}

\vskip 0.3in
]

\begin{abstract}
It is critical in many applications to understand what features are important for a model, and why individual predictions were made. For tree ensemble methods these questions are usually answered by attributing importance values to input features, either globally or for a single prediction. Here we show that current feature attribution methods are inconsistent, which means changing the model to rely more on a given feature can actually decrease the importance assigned to that feature.
To address this problem we develop fast exact solutions for SHAP (\underline{SH}apley \underline{A}dditive ex\underline{P}lanation) values, which were recently shown to be the unique additive feature attribution method based on conditional expectations that is both consistent and locally accurate. We integrate these improvements into the latest version of XGBoost, demonstrate the inconsistencies of current methods, and show how using SHAP values results in significantly improved supervised clustering performance.
Feature importance values are a key part of understanding widely used models such as gradient boosting trees and random forests. We believe our work improves on the state-of-the-art in important ways, and so impacts any current user of tree ensemble methods.
\end{abstract}  

\section{Introduction}

Understanding why a model made a prediction is important for trust, actionability, accountability, debugging, and many other common tasks. To understand predictions from tree ensemble methods, such as gradient boosting trees or random forests, importance values are typically attributed to each input feature. These importance values can be computed either for a single prediction, or an entire dataset to explain a model's overall behavior.

Concerningly, current feature attribution methods for tree ensembles are {\it inconsistent}, meaning they can assign higher importance to features with a lower impact on the model's output. This inconsistency effects a very large number of users, since tree ensemble methods are widely applied in research and industry.

Here we show that by connecting tree ensemble feature attribution methods with the recently defined class of {\it additive feature attribution methods} \cite{lundberg2017unified} we can motivate the use of SHapley Additive exPlanation (SHAP) values as the only possible consistent feature attribution method with desirable properties.

SHAP values are theoretically optimal but can be challenging to compute. To address this we derive exact algorithms for tree ensemble methods that reduce the computational complexity of computing SHAP values from exponential to $O(T LD^2)$ where $T$ is the number of trees, $L$ is the maximum number of leaves in any tree, and $D$ is the maximum depth of any tree. By integrating this new algorithm into XGBoost, a popular tree ensemble package, we demonstrate performance that enables predictions from models with thousands of trees, and hundreds of inputs, to be explained in a fraction of a second.

In what follows we first discuss the inconsistencies of current feature attribution methods as implemented in popular tree enemble software packages (Section \ref{sec:current}). We then introduce SHAP values as the only possible consistent attributions (Section \ref{sec:shap}), and present Tree SHAP as a high speed algorithm for estimating SHAP values of tree ensembles (Section \ref{sec:tree_shap}). Finally, we use a supervised clustering task to compare SHAP values with previous feature attribution methods (Section \ref{sec:experiments}).

\begin{figure*}
  \centering
  \includegraphics[width=0.7\textwidth]{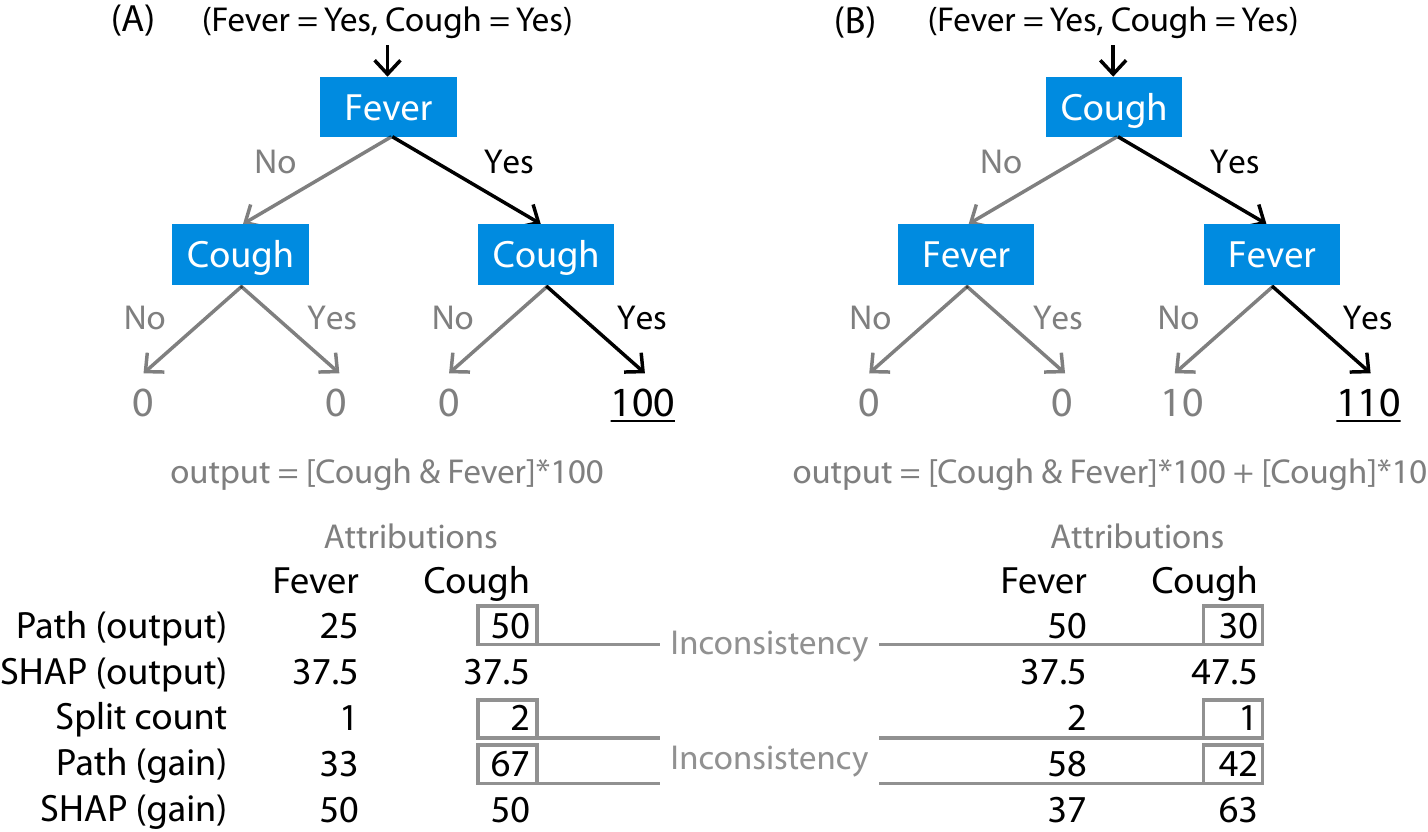}
  \caption{Two tree models meant to demonstrate the inconsistencies of current feature attribution methods. The Cough feature has a larger impact on tree B, but is assigned less importance by all three standard methods. The ``output" attributions explain the difference between the expected value of the model output and the current output. The ``gain" represents the change in the mean squared error over the whole dataset between when no features are used and all features are used. All calculations assume a dataset (typically a training dataset) perfectly matching the model and evenly spread among all leaves. Section 2 describes the standard ``path" methods, while Section \ref{sec:shap} describes the SHAP values and their interpretation.}
  \label{fig:and_trees}
\end{figure*}

\section{Current feature attributions are inconsistent}
\label{sec:current}

Tree ensemble implementations in popular packages such as XGBoost \cite{chen2016xgboost}, scikit-learn \cite{pedregosa2011scikit}, and the {\it gbm} R package \cite{ridgeway2010generalized}, allow a user compute a measure of feature importance. These values are meant to summarize a complicated ensemble model and provide insight into what features drive the model's prediction. Unfortunately the standard feature importance values provided by all of these packages are inconsistent, this means that a model can change such that it relies more on a given feature, yet the importance assigned to that feature decreases (Figure \ref{fig:and_trees}).

For the above packages, when feature importance values are calculated for an entire dataset they are by default based on the reduction of loss (termed ``gain") contributed by each split in each tree of the ensemble. Feature importances are then defined as the sum of the gains of all splits for a given feature as described in \citeauthor{friedman2001elements} \cite{breiman1984classification,friedman2001elements}.

Methods computing feature importance values for a single prediction are less established, and of the above packages, only the most recent version of XGBoost supports these calculations natively. The method used by XGBoost \cite{path_blog} is similar to the classical dataset level feature importance calculation, but instead of measuring the reduction of loss it measures the change in the model's output.

Both current feature attribution methods described above only consider the effect of splits along the decision path, so we will term them {\it path} methods. Figure \ref{fig:and_trees} shows the result of applying both these methods to two simple regression trees. For the gain calculations we assume equal coverage of each of the four tree leaves, and perfect regression accuracy. In other words, an equal number of dataset points fall in each leaf, and the label of those points is exactly equal to the prediction of the leaf. The tree in Figure \ref{fig:and_trees}A represents a simple AND function, while the tree in Figure \ref{fig:and_trees}B represents the same AND function but with an additional increase in predicted value when Cough is ``Yes".

The point of Figure \ref{fig:and_trees} is to compare feature attributions between A and B, where it is clear that Cough has a larger impact on the model in B than the model in A. As highlighted below each tree, we can see that current path methods (as well as the simple split count metric) are inconsistent because they allocate less importance to Cough in B, even though Cough has a larger impact on the output of the tree in B. The ``output" task explains the change in model output from the expected value to the current predicted value given Fever and Cough. The ``gain" explains the reduction in mean squared error contributed by each feature (assuming a dataset as described in the previous paragraph). In contrast to current approaches, the SHAP values (described below) are consistent, even when the order in which features appear in the tree changes.

\section{SHAP values are the only consistent feature attributions}
\label{sec:shap}

\begin{figure*}
  \centering
  \includegraphics[width=1.0\textwidth]{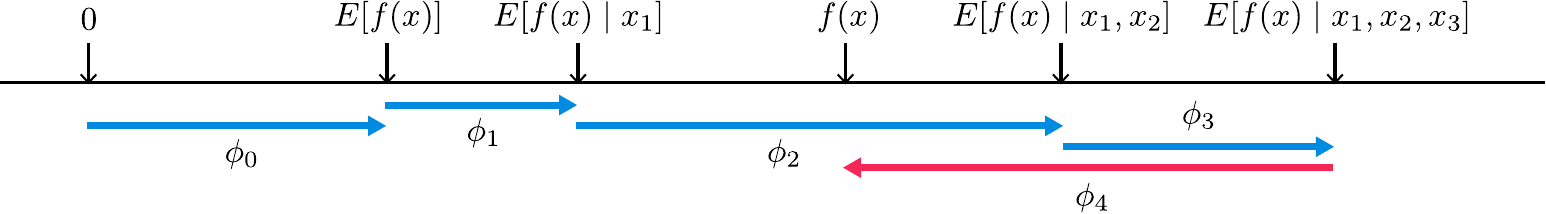}
  \caption{SHAP (\underline{SH}apley \underline{A}dditive ex\underline{P}lanation) values explain the output of a function as a sum of the effects $\phi_i$ of each feature being introduced into a conditional expectation. Importantly, for non-linear functions the order in which features are introduced matters, so SHAP averages over all possible orderings. Proofs from game theory show this is the only possible consistent and locally accurate approach. In contrast, standard path methods for tree ensembles (Section \ref{sec:current}) are similar to using a single ordering defined by a tree's decision path.}
  \label{fig:number_line}
\end{figure*}

It was recently noted that many current methods for interpreting machine learning model predictions fall into the class of {\it additive feature attribution methods} \cite{lundberg2017unified}. This class covers all methods that explain a model's output as a sum of real values attributed to each input feature.

\begin{definition}
\label{def:additive}
{\bf Additive feature attribution methods} have an explanation model that is a linear function of binary variables:
\begin{equation}
\label{eq:additive_fa}
g(z') = \phi_0 + \sum_{i = 1}^M \phi_i z_i',
\end{equation}
where $z' \in \{0,1\}^M$, $M$ is the number of input features, and $\phi_i \in \mathbb{R}$.
\end{definition}

The $z'_i$ variables typically represent a feature being observed ($z'_i = 1$) or unknown ($z'_i = 0$), and the $\phi_i$'s are the feature attribution values.

As previously described in \citeauthor{lundberg2017unified}, an important attribute of the class of additive feature attribution methods is that there is a single unique solution in this class with three desirable properties: {\it local accuracy}, {\it missingness}, and {\it consistency} \cite{lundberg2017unified}. Local accuracy states that the sum of the feature attributions is equal to the output of the function we are seeking to explain. Missingness states that features that are already missing (such that $z'_i = 0$) are attributed no importance. Consistency states that changing a model so a feature has a larger impact on the model, will never decrease the attribution assigned to that feature.

In order to evaluate the effect missing features have on a model $f$, it is necessary to define a mapping $h_x$ that maps between the original function input space and the binary pattern of missing features represented by $z'$. Given such a mapping we can evaluate $f(h_x^{-1}(z'))$ and so calculate the effect of observing or not observing a feature (by setting $z'_i = 1$ or $z'_i = 0$).

SHAP values define $f_x(S) = f(h_x^{-1}(z')) = E[f(x) \mid x_S]$ where $S$ is the set of non-zero indexes in $z'$ (Figure \ref{fig:number_line}), and then use the classic Shapley values from game theory to attribute $\phi_i$ values to each feature:

\begin{equation}
\label{eq:shapley}
\phi_i = \sum_{S \subseteq N \setminus \{i\}} \frac{|S|!(M - |S|! -1)}{M!} \left [ f_x(S \cup \{i\}) - f_x(S) \right ]
\end{equation}

\noindent where $N$ is the set of all input features.

The SHAP values are the only possible consistent, locally accurate method that obeys the missingness property and uses conditional dependence to measure missingness \cite{lundberg2017unified}. This is strong motivation to use SHAP values for tree ensemble feature attribution, particularly since current tree ensemble feature attribution methods already obey all of these properties except consistency. This means that SHAP values provide a strict theoretical improvement over existing approaches by eliminating the unintuitive consistency problems shown in Figure \ref{fig:and_trees}.

\section{Tree SHAP: Fast SHAP value computation for decision trees}
\label{sec:tree_shap}

Despite the compelling theoretical advantages of SHAP values, their practical use is hindered by two problems:

\begin{enumerate}
\item The challenge of estimating $E[f(x) \mid x_S]$ efficiently.
\item The exponential complexity of Equation \ref{eq:shapley}.
\end{enumerate}

Here we focus on tree models and propose fast SHAP value estimation methods specific to trees and ensembles of trees. We start by defining a straightforward, but slow, algorithm in Section \ref{sec:shap_direct}, then present the much faster and more complex Tree SHAP algorithm in Section \ref{sec:shap_fast}.

\subsection{Estimating SHAP values directly in $O(TL2^M)$ time}
\label{sec:shap_direct}

If we ignore computational complexity then we can compute the SHAP values for a decision tree by estimating $E[f(x) \mid x_S]$ and then using Equation \ref{eq:shapley} where $f_x(S) = E[f(x) \mid x_S]$. For a tree model $E[f(x) \mid x_S]$ can be estimated recursively using Algorithm \ref{alg:exp_value}, where $v$ is a vector of node values, which takes the value $internal$ for internal nodes. The vectors $a$ and $b$ represent the left and right node indexes for each internal node. The vector $t$ contains the thresholds for each internal node, and $d$ is a vector of indexes of the features used for splitting in internal nodes. The vector $r$ represents the cover of each node (how many data samples fall in that subtree).

\begin{algorithm}
\caption{Estimating $E[f(x) \mid x_S]$ \label{alg:exp_value}}
\begin{algorithmic}
\Procedure{EXPVALUE}{$x$, $S$, $\text{tree} = \{v, a, b, t, r, d\}$}
\Procedure{G}{$j$, $w$}
  \If{$v_j \ne internal$}
    \State \Return{$w \cdot v_j$}
  \Else
    \If{$d_j \in S$}
      \State \Return $x_{d_j} \le t_j~?~\text{\Call{G}{$a_j$, $w$}} : \text{\Call{G}{$b_j$, $w$}}$
    \Else
    \State \Return \Call{G}{$a_j$, $w r_{a_j} / r_j$} + \Call{G}{$a_j$, $w r_{a_j} / r_j$}
    \EndIf
    
  \EndIf
\EndProcedure
\State \Return \Call{G}{$1$, $1$}
\EndProcedure
\end{algorithmic}
\end{algorithm}

\subsection{Estimating SHAP values in $O(TLD^2)$ time}
\label{sec:shap_fast}

Here we propose a novel algorithm to calculate the same values as in Section \ref{sec:shap_direct}, but in polynomial time instead of exponential time. Specifically, we propose an algorithm that runs in $O(TL \log^2 L)$ for balanced trees, and $O(TLD^2)$ for unbalanced trees.

The general idea of the polynomial time algorithm is to recursively keep track of what proportion of all possible subsets flow down into each of the leaves of the tree. 
This is similar to running Algorithm \ref{alg:exp_value} simultaneously for all $2^M$ subsets $S$ in Equation \ref{eq:shapley}. It may seem reasonable to simply keep track of how many subsets (weighted by the cover splitting of Algorithm \ref{alg:exp_value}) pass down each branch of the tree. However, this combines subsets of different sizes and so prevents the proper weighting of these subsets, since the weights in Equation \ref{eq:shapley} depend on $|S|$. To address this we keep track of each possible subset size during the recursion. The {\it EXTEND} method in Algorithm \ref{alg:tree_shap} grows all these subsets according to given fraction of ones and zeros, while the {\it UNWIND} method reverses this process. The {\it EXTEND} method is used as we descend the tree. The {\it UNWIND} method is used to undo previous extensions when we split on the same feature twice, and to undo each extension of the path inside a leaf to correctly compute the weights for each feature in the path.

In Algorithm \ref{alg:tree_shap}, $m$ is the path of unique features we have split on so far, and contains four attributes: $d$ the feature index, $z$ the fraction of ``zero" paths (where this feature is not in the set $S$) that flow through this branch, $o$ the fraction of ``one" paths (where this feature is in the set $S$) that flow through this branch, and $w$ which is used to hold the proportion of sets of a given cardinally that are present. We use the dot notation to access these members, and for the whole vector $m.d$ represents a vector of all the feature indexes. (For code see https://github.com/slundberg/shap)

\begin{algorithm}
\caption{Tree SHAP \label{alg:tree_shap}}
\begin{algorithmic}
\Procedure{TS}{$x$, $\text{tree} = \{v, a, b, t, r, d\}$}
\State $\phi = \text{array of $len(x)$ zeros}$
\Procedure{RECURSE}{$j$, $m$, $p_z$, $p_o$, $p_i$}
  \State $m =~$\Call{EXTEND}{$m$, $p_z$, $p_o$, $p_i$}
  \If{$v_j \ne internal$}
    \For{$i \gets 2 \textrm{ to } len(m)$}
      \State $w = sum(\text{\Call{UNWIND}{$m$, $i$}}.w)$
      \State $\phi_{m_i} = \phi_{m_i} + w(m_i.o - m_i.z) v_j$
    \EndFor
  \Else
    \State $h,c = x_{d_j} \le t_j~?~(a_j,b_j) : (b_j,a_j)$
    \State $i_z = i_o = 1$
    \State $k = \text{\Call{FINDFIRST}{$m.d$, $d_j$}}$
    \If{$k \ne \text{nothing}$}
      \State $i_z,i_o = (m_k.z,m_k.o)$
      \State $m = \text{\Call{UNWIND}{$m$, $k$}}$
    \EndIf
    \State \Call{RECURSE}{$h$, $m$, $i_z r_h/r_j$, $i_o$, $d_j$}
    \State \Call{RECURSE}{$c$, $m$, $i_z r_c/r_j$, $0$, $d_j$}
  \EndIf
\EndProcedure
\Procedure{EXTEND}{$m$, $p_z$, $p_o$, $p_i$}
\State $l = len(m)+1$
\State $m = copy(m)$
\State $m_{l+1}.(d,z,o,w) = (p_i, p_z, p_o, l = 0~?~1:0)$
\For{$i \gets l-1 \textrm{ to } 1$}
  \State $m_{i+1}.w = m_{i+1}.w + p_o m_i.w(i/l)$
  \State $m_i.w =  p_z m_i.w[(l-i)/l]$
\EndFor
\State \Return m
\EndProcedure
\Procedure{UNWIND}{$m$, $i$}
\State $l = len(m)$
\State $n = m_l.w$
\State $m = copy(m_{1...l-1})$
\For{$j \gets l-1 \textrm{ to } 1$}
  \If{$m_i.o \ne 0$}
    \State $t = m_j.w$
    \State $m_j.w = n \cdot l/(j \cdot m_i.o)$
    \State $n = t - m_j.w \cdot m_i.z ((l-j)/l)$
  \Else
    \State $m_j.w = (m_j.w \cdot l)/(m_i.z (l-j))$
  \EndIf
\EndFor
\For{$j \gets i \textrm{ to } l-1$}
  \State $m_j.(d,z,o) = m_{j+1}.(d,z,o)$
\EndFor
\State \Return m
\EndProcedure
\State \Call{RECURSE}{$1$, $[]$, $1$, $1$, $0$}
\State \Return $\phi$
\EndProcedure
\end{algorithmic}
\end{algorithm}


\begin{figure*}
  \centering
  \includegraphics[width=1.0\textwidth]{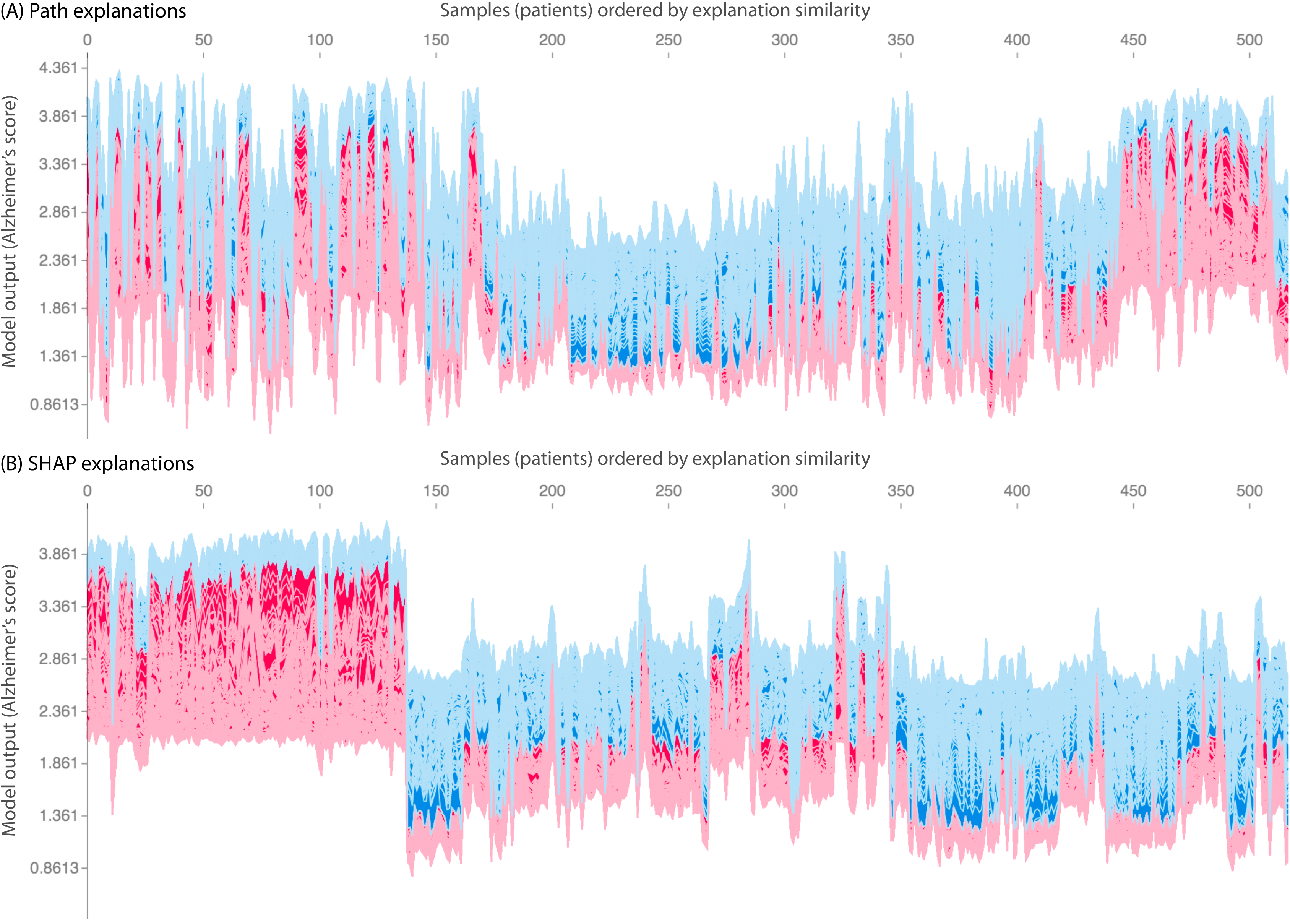}
  \caption{SHAP feature attributions produce better clusters than standard path attributions for supervised clustering of 518 participants in an Alzheimer's research study. An XGBoost model with 300 trees of max depth six was trained on 200 gene expression module features using a shrinkage factor of $\eta = 0.01$. This model was then used to predict the CERAD cognitive score of each participant. Each prediction was explained, and then clustered using hierarchical agglomerative clustering (imagine a dendrogram joining the samples above each plot). Red feature attributions push the score higher, while blue feature attributions push the score lower. A) The clusters formed with standard ``path" explanations from XGBoost. B) Clusters using our Tree SHAP XGBoost implementation.}
  \label{fig:clustering}
\end{figure*}
\newpage
\section{Supervised clustering experiments}
\label{sec:experiments}

One intriguing use for prediction level feature attributions is what we term ``supervised clustering", where instead of using an unsupervised clustering method directly on the data features, you run clustering on the feature attributions \cite{lundberg2016unexpected}.

Supervised clustering naturally handles one of the most challenging problems in unsupervised clustering: determining feature weightings (or equivalently, determining a distance metric). Many times we want to cluster data using features with very different units. 
Features may be in dollars, meters, unit-less scores, etc. but whenever we use them as dimensions in a single multidimensional space it forces any distance metric to compare the relative importance of a change in different units (such as dollars vs. meters). Even if all our inputs are in the same units, often some features are more important than others. Supervised clustering uses feature attributions to naturally convert all the input features into values with the same units as the model output. This means that a unit change in any of the feature attributions is comparable to a unit change in any other feature attribution. It also means that fluctuations in the feature values only effect the clustering if those fluctuations have an impact on the outcome of interest.

Here we compare feature attribution methods by applying supervised clustering to disease sub-typing, an area where unsupervised clustering has contributed to important discoveries. The goal of disease sub-typing is to identify subgroups of patients that have similar mechanisms of disease (similar reasons they are sick). Here we consider Alzheimer's disease where the predicted outcome is the CERAD cognitive score \cite{mirra1991consortium}, and the features are gene expression modules \cite{celik2014efficient}.

By representing the positive feature attributions as red bars and the negative feature attributions as blue bars (as in Figure \ref{fig:number_line}), we can stack them against each other to visually represent the model output as their sum. Figure \ref{fig:clustering} does this vertically for each participant. The explanations for each participant are then stacked horizontally according the leaf order of a hierarchical clustering. This groups participants with similar predicted outcomes and similar reasons for that predicted outcome together. The clearer structure in Figure \ref{fig:clustering}B indicates the SHAP values are better feature attributions, not only theoretically, but also practically.

\begin{figure}
  \centering
  \includegraphics[width=0.4\textwidth]{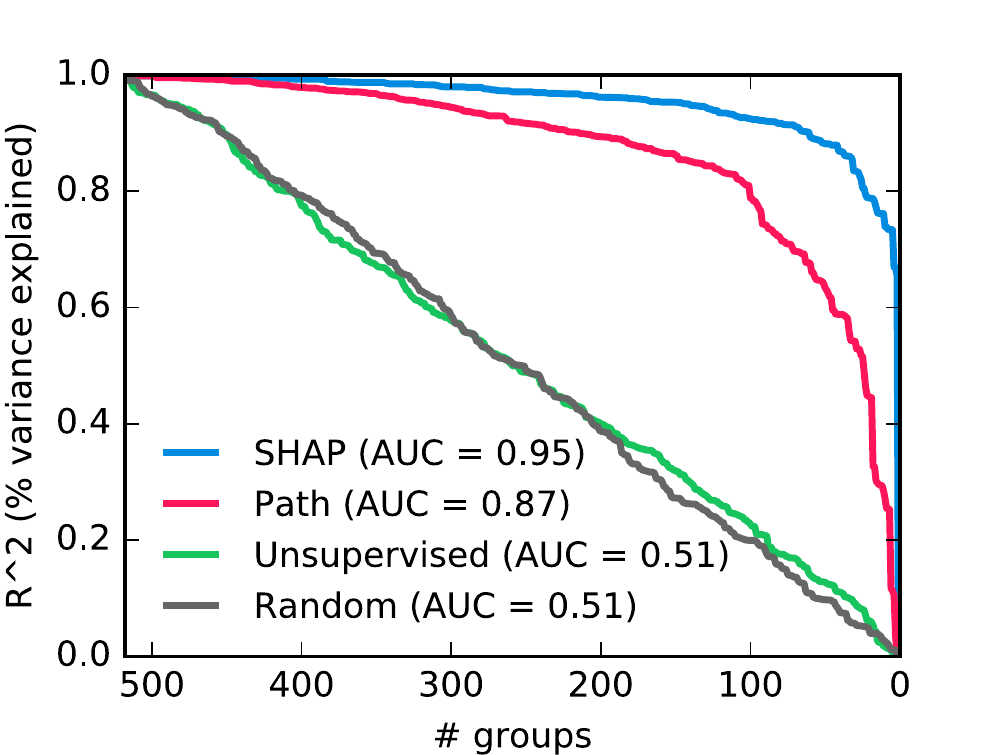}
  \caption{A quantitative performance measure of the clusterings shown in Figure \ref{fig:clustering}. If all 518 samples are placed in their own group, and each group predicts the mean value of the group, then the $R^2$ value (the proportion of outcome variance explained) will be $1$. If groups are then merged one-by-one the $R^2$ will decline until when there is only a single group it will be $0$. Hierarchical clusterings that well separate the outcome value will retain a high $R^2$ longer during the merging process. Here unsupervised clustering did no better than random, supervised clustering with the XGBoost ``path" method did significantly better, and SHAP values significantly better still.}
  \label{fig:auc}
\end{figure}

The improvement in clustering performance seen in Figure \ref{fig:clustering} can be quantified by examining how well each clustering explains the variance of the CERAD score outcome. Since hierarchical clusterings encode many possible groupings, we plot in Figure \ref{fig:auc} the change in the $R^2$ value as the number of groups shrinks from one group per sample ($R^2 = 1$), to a single group ($R^2 = 0$).

\section{Conclusion}
Here we have shown that classic feature attribution methods for tree ensembles are inconsistent, meaning they can assign less importance to a feature when the true effect of that feature increases. In contrast, SHAP values were shown to be the unique way to consistently attribute feature importance. By deriving fast algorithms for SHAP values and integrating them with XGBoost, we make them a practical replacement for previous methods. Future directions include deriving fast dataset-level  SHAP algorithms for gain (as opposed to the instance-level algorithm presented here), and integrating SHAP value algorithms into the released versions of common packages.

\newpage
\subsection*{Acknowledgments}
We would like to thank Gabriel Erion for suggestions that lead to a simplified algorithm, as well as Jacob Schreiber and Naozumi Hiranuma for providing helpful input.  

\bibliography{shap_icml}
\bibliographystyle{icml2017}

\end{document}